\pdfoutput=1

\documentclass[11pt]{article}

\usepackage[final]{acl}
\usepackage{amsmath}
\usepackage{times}
\usepackage{latexsym}
\usepackage{makecell}
\usepackage{times}
\usepackage{marvosym}
\usepackage{latexsym}
\usepackage{booktabs}
\usepackage{multirow}
\usepackage{graphicx} 
\usepackage{subcaption}
\usepackage{makecell}
\usepackage{amssymb}  
\usepackage{graphicx}
\usepackage{hyperref}

\usepackage[T1]{fontenc}

\usepackage[utf8]{inputenc}

\usepackage{microtype}

\usepackage{inconsolata}

\usepackage{graphicx}

\title{An Effective Deployment of Diffusion LM for Data Augmentation in Low-Resource Sentiment Classification}

\author{Zhuowei Chen${}^{1}$, Lianxi Wang${}^{1,2\footnotemark[2]}$,\\ \textbf{Yuben Wu${}^{1}$, Xinfeng Liao${}^{1}$, Yujia Tian${}^{1}$, Junyang Zhong${}^{1}$} \\
       ${}^1$Guangdong University of Foreign Studies, Guangzhou, China.\\
       ${}^2$Guangzhou Key Laboratory of Multilingual Intelligent Processing, Guangzhou, China.\\
       \texttt{wanglianxi@gdufs.edu.cn}
       }

\begin{document}
\maketitle

\renewcommand{\thefootnote}{\fnsymbol{footnote}}
\footnotetext[2]{Corresponding author.}
\renewcommand{\thefootnote}{\arabic{footnote}}



\begin{abstract}


Sentiment classification (SC) often suffers from low-resource challenges such as domain-specific contexts, imbalanced label distributions, and few-shot scenarios. The potential of the diffusion language model (LM) for textual data augmentation (DA) remains unexplored, moreover, textual DA methods struggle to balance the diversity and consistency of new samples. Most DA methods either perform logical modifications or rephrase less important tokens in the original sequence with the language model. In the context of SC, strong emotional tokens could act critically on the sentiment of the whole sequence. Therefore, contrary to rephrasing less important context, we propose DiffusionCLS to leverage a diffusion LM to capture in-domain knowledge and generate pseudo samples by reconstructing strong label-related tokens. This approach ensures a balance between consistency and diversity, avoiding the introduction of noise and augmenting crucial features of datasets. DiffusionCLS also comprises a Noise-Resistant Training objective to help the model generalize. Experiments demonstrate the effectiveness of our method in various low-resource scenarios including domain-specific and domain-general problems. Ablation studies confirm the effectiveness of our framework's modules, and visualization studies highlight optimal deployment conditions, reinforcing our conclusions. 
\end{abstract}

\section{Introduction}

Sentiment classification is a crucial application of text classification (TC) in Natural Language Processing (NLP) and can play a crucial role in multiple areas. However, NLP applications in domain-specific scenarios, such as disasters and pandemics, often meet with low-resource conditions, especially domain-specific problems, imbalance data distribution, and data deficiency \cite{sedinkina2022domain,lakshmi2023classification,nabil2023bangla,gatto2023not}.  Recently, the birth of pre-trained language models (PLMs) and large language models (LLMs) have advanced the NLP field, giving birth to numerous downstream models based on them. On the one hand, these PLMs take the models to a new height of performance, on the other hand, since these models are highly data-hungry, they struggle to perform satisfactorily on most tasks under noisy, data-sparse and low-resource conditions \cite{patwa-etal-2024-enhancing-low,chen2023distantly,wang-etal-2024-enhancing-hindi,yu-etal-2023-cross}.

\begin{table}[h!]
\scalebox{0.72}{
\centering
    \begin{tabular}{rrl}
    \toprule
    \multicolumn{1}{c|}{Method} & \multicolumn{1}{c|}{Type} & Textual Sample \\
    \midrule
    \multicolumn{1}{c|}{\multirow{2}[5]{*}{\makecell{Other\\CTR Methods\\(GENIUS)}}} & \multicolumn{1}{c|}{Cor.} & \makecell{\textit{[sad]} [M] the traffic [M] a nightmare.\\{[M][M][M]}frustrating.} \\
\cmidrule{2-3}    \multicolumn{1}{c|}{} & \multicolumn{1}{c|}{Gen.} & \makecell{\textit{[sad]} \textcolor[rgb]{ .753,  0,  0}{Navigating} the traffic \textcolor[rgb]{ .753,  0,  0}{was literally}\\ a nightmare. \textcolor[rgb]{ .753,  0,  0}{Truly} frustrating.} \\
    \midrule
    \multicolumn{1}{c|}{\multirow{2}[5]{*}{\makecell{Diffusion-\\CLS\\(ours)}}} & \multicolumn{1}{c|}{Cor.} & \makecell{\textit{[sad]} Today, the [M] was [M][M].\\It was [M][M].} \\
\cmidrule{2-3}    \multicolumn{1}{c|}{} & \multicolumn{1}{c|}{Gen.} & \makecell{\textit{[sad]} Today, the \textcolor[rgb]{ 0,  .69,  .314}{journey} was a \textcolor[rgb]{ 0,  .69,  .314}{disaster}.\\It was \textcolor[rgb]{ 0,  .69,  .314}{utterly} \textcolor[rgb]{ 0,  .69,  .314}{chaotic}.} \\
    \midrule
    \multicolumn{1}{r}{Original Text:} & \multicolumn{2}{l}{\makecell{\textcolor[rgb]{ .753,  0,  0}{Today,} the \textcolor[rgb]{ 0,  .69,  .314}{traffic} \textcolor[rgb]{ .753,  0,  0}{was} a \textcolor[rgb]{ 0,  .69,  .314}{nightmare}.\\\textcolor[rgb]{ .753,  0,  0}{It was really} \textcolor[rgb]{ 0,  .69,  .314}{frustrating.}}} \\
    \bottomrule
    \end{tabular}%

}
  \caption{Examples of CTR methods. Most CTR methods rephrase minor tokens while DiffusionCLS reconstructs strong label-related tokens. Cor. and Gen. denotes the corrupted sequence and generated sequence respectively.}

  \label{tab:example}%
\end{table}%

To address these challenges, one effective approach is data augmentation (DA), which enriches the diversity of the dataset without explicitly collecting new data \cite{feng2021survey}. Classic rule-based DA methods employ logical modifications to obtain pseudo samples, such as EDA \cite{wei-zou-2019-eda}, and AEDA \cite{karimi-etal-2021-aeda-easier}. Model-based DA methods develop rapidly as the transformer architecture dominates the NLP field, most of these methods execute DA through corrupt-then-reconstruct (CTR), as examples shown in Table \ref{tab:example}. Namely, masked language model (MLM) \cite{wu2019conditional,kumar2020data}, and GENIUS \cite{guo2022genius} which applies BART as the sample generator. Also, \citet{anabytavor2019data} proposed LAMBADA, which finetunes GPT-2 and generates pseudo samples with label prompts.

However, these methods often struggle with domain-specific tasks and uneven label distributions. Some methods generate samples solely relying on pre-trained knowledge, like GENIUS. The other though finetuned on the downstream dataset, these methods generate samples only conditioned on the label itself, such as LAMBADA, leading to strong label inconsistency, especially in data-sparse settings. Also, most CTR methods focus on replacing minor tokens in sequences but keeping the crucial tokens stationary to generate high-quality pseudo samples. 

In contrast, we corrupt the most label-related tokens first and reconstruct the whole sentence conditioned on the context and label prompt, as shown in Table \ref{tab:example}, to diversify the key label-related tokens rather than less important contexts. This approach not only augments sample diversity but also upholds consistency through selective masking. Inspired by DiffusionBERT \cite{he-etal-2023-diffusionbert}, which is designed to recover the most informative tokens from those with less informatics, we propose DiffusionCLS. Additionally, building upon the findings of \citet{guo2022genius}, we further introduce consistency and diversity as crucial elements for quality of samples. High-quality pseudo samples must align with their labels and domain contexts, minimizing noise introduction. Integrating these samples enhances dataset diversity, thereby positively impacting the model performance.

DiffusionCLS initially finetunes PLM with a diffusion objective, functioning as a sample generator, followed by training the TC model in a noise-resistant manner. By fine-tuning the diffusion LM, we can then input original samples with their crucial tokens corrupted and use the label as a generation prompt to get new samples. This method diversifies the original dataset by replacing strong label-related tokens and also steers the model towards producing high-quality pseudo samples that comply with the diversity-consistency rule. Also, experimental codes have been released on GitHub\footnote{https://github.com/JohnnyChanV/DiffusionCLS}.

The major contributions of this paper can be summarized as follows:
\begin{itemize}
    \item  We propose DiffusionCLS, which comprises a diffusion LM-based data augmentation module for SC, generating diverse but consistent pseudo samples by substituting diverse strong label-related contexts.
    
    \item Designed and integrated a noise-resistant training method within the proposed DiffusionCLS, which significantly improves the SC model's performance with pseudo samples. 
    
    \item Comprehensive experiments on domain-specific and multilingual datasets validate DiffusionCLS's superior performance in SC tasks. Detailed ablation studies highlights the effectiveness of its individual modules.

    \item A visualization study is conducted to discuss the diversity-consistency trade-off, which further validates the effectiveness of DiffusionCLS.
\end{itemize}

\begin{figure*}
    \centering
    \includegraphics[width=1\textwidth]{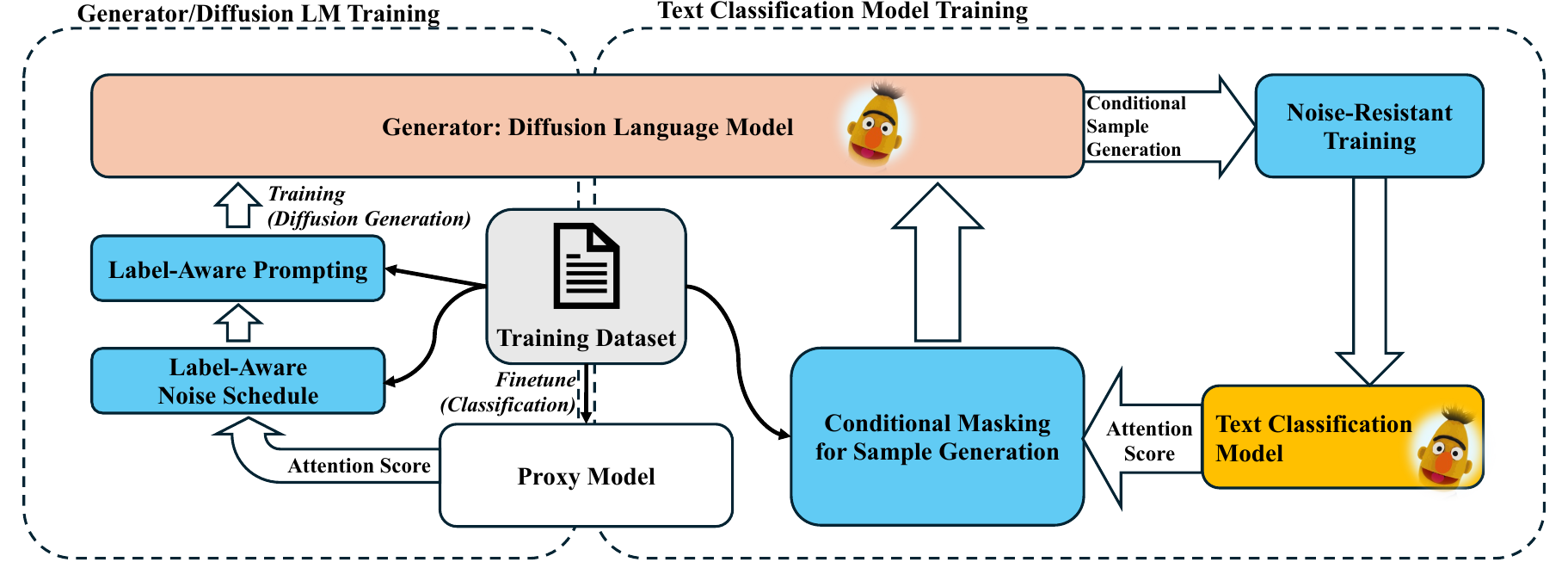}
    \caption{Overview of the proposed method. DiffusionCLS comprises four core components: Label-Aware Noise Schedule, Label-Aware Prompting, Conditional Sample Generation, and Noise-Resistant Training.}
    \label{fig:overview}
\end{figure*}

\section{Related Work}

\subsection{Low-Resource Text Classification}



Motivated by the observation that data is often scarce in specific domains or emergent application scenarios, low-resource TC \cite{chen-etal-2018-adversarial} has recently attracted considerable attention. Low-resource TC involves effectively categorizing text in scenarios where data is scarce or limited. 
\citet{goudjil2018novel} and \citet{tan2019out} have explored several methods for low-resource TC, which mainly involve traditional machine learning techniques to increase data quantity and diversity.
 
Recently, since the studies by \citet{lan2019albert} and \citet{sun2020ernie} demonstrated the impressive performance of PLMs across various NLP tasks, a significant amount of work has leaned towards using PLMs to address low-resource TC problems \cite{wen2023augmenting,ogueji2021small,liu2019roberta,devlin2018bert}. However, PLMs requires amounts of annotated samples for finetuning, data-sparce significantly impacts models' performances and DA could mitigate such problems.

\subsection{Textual Data Augmentation}



To address low-resource challenges, various data augmentation methods have been proposed, including Easy-Data-Augmentation (EDA) \cite{wei-zou-2019-eda}, Back-Translation (BT) \cite{shleifer2019low}, and CBERT \cite{wu2019conditional}. However, these methods, relying on logical replacements and external knowledge, often introduce out-domain knowledge and domain inconsistency. Moreover, these methods focus only on a specific original input, resulting in limited diversity.

Another type of data augmentation method includes representation augmentation approaches. These methods generate pseudo-representation vectors by interpolating or perturbing the representations of original samples. For instance, \citet{zhang2017mixup} proposed the groundbreaking technique known as mixup, and \citet{chen2023adversarial} recently proposed AWD, an advanced approach in textual DA.

Recent advancements in generative models have led to research on GPT-based paraphrasing data augmentation methods, such as LAMBADA \cite{anabytavor2019data}, which fine-tuned GPT-2 model to generate new samples. However, LAMBADA generates new samples based solely on specific labels, neglecting information from the original samples. Another research direction involves not fine-tuning PLMs but combining the language modeling capability of pretrained models with the generative diversity of diffusion models \cite{he-etal-2023-diffusionbert}, which significantly improves the capability of the generative encoder, i.e., MLM.

Since diffusion LMs can generate new sequences from masked original sequences, which matches the goal of retaining key information and rephrasing secondary information in generative data augmentation. Therefore, on top of diffusion LM, we propose DiffusionCLS, simultaneously considering label and domain consistency and generating pseudo samples by partially paraphrasing strong label-related tokens. Extensive experiments verify the effectiveness of our method and hopefully be extended to numerous NLP tasks.


\section{Methodology}

Sentiment classification models often overfit and lack generalization due to sample deficiency. To address this, we propose DiffusionCLS, consisting of Label-Aware Noise Schedule, Label-Aware Prompting, Conditional Sample Generation, and Noise-Resistant Training. A diffusion LM-based sample generator is integrated to generate new samples from the original dataset, enhancing TC model performance.

Figure \ref{fig:overview} illustrates DiffusionCLS. The diffusion LM-based sample generator generates new samples for data augmentation, while the TC model is trained for the specific task. Label-Aware Prompting and Label-Aware Noise Schedule are crucial for training the sample generator, and Conditional Sample Generation and Noise-Resistant Training contribute to the training of the TC model.

\subsection{Sample Generator}
To generate usable samples for further TC model training, there are two crucial rules of success to satisfy, diversity and consistency. Therefore, we expect the generated samples to be as diverse as possible with consistency to the TC label and original domain simultaneously. However, higher diversity also leads to a higher difficulty in maintaining consistency. 

As \citet{he-etal-2023-diffusionbert} excavated the potential of combining diffusion models with LMs for sequence generation, we built the sample generator from the discrete diffusion model scratch. Precisely, we design the Label-Aware Noise Schedule for the diffusion LM, which helps the model to generate diverse and consistent samples. Additionally, we integrate Label-Aware Prompting into the training regime, enabling the model to grasp label-specific knowledge, subsequently serving as the guiding condition for sample generation. These two modules help the generator to surpass the diversity-consistency challenge and excel in performance.


\subsubsection{Label-Aware Noise Schedule}
A proper algorithm of noise schedule could guide the diffusion LM to capture more accurate semantic relations. Moreover, the effectiveness of time-agnostic decoding has been demonstrated, indicating that incorporating implicit time information in the noise schedule process is effective \cite{ho2020denoising,nichol2021improved,he-etal-2023-diffusionbert}. Since the generated samples are also expected to stay consistent with the TC label and the original domain, we proposed Label-Aware Noise Schedule. 

The Label-Aware Noise Schedule begins by integrating a proxy model that has been fine-tuned for the TC task. This proxy model allows for the determination of the importance of each token in the TC process, quantified through attention scores between the [CLS] token and other tokens, which are derived from the last layer of proxy model and calculated as follows.

\begin{equation}
    w_i = \frac{1}{H}\sum_{h=1}^{H} {s_i^h},
        \label{tokenWCal}
\end{equation}
where $s_i^h$ represents the $i$-th token attention score in the $h$-th attention head, and $w_i$ denotes the weight measuring the importance of the $i$-th token.

Motivated by \citet{he-etal-2023-diffusionbert}'s DiffusionBERT, we incorporates absorbing state in the LM noise schedule. In our method, during the masking transition procedure, each token in the sequence remains unchanged or transitions to [MASK] with a certain probability. The transition probability of token $i$ at step $t$ can be denoted as:
\begin{equation}
	q_t^i = 1 - \frac{t}{T} - \lambda \cdot S(t) \cdot w_i,
\end{equation}

\begin{equation}
    S(t) = \mathrm{sin} \frac{t\pi}{T},
\end{equation}
where $q_t^i$ represents the probability that a token is being masked, and $T$ denotes the total step number. $\lambda$ is introduced to control the impact of $w_i$, as a hyper-parameter. 

\begin{figure}[h!]
\centering
\includegraphics[width=.38\textwidth]{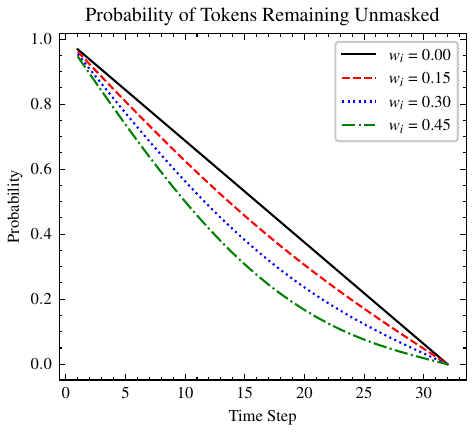}
\caption{The probability of a token remaining unmasked, with $\lambda$ set to 0.5.}
\label{probMask}
    \vspace{-8pt}
\end{figure}

By introducing strong label-related $w_i$, the diffusion model is guided to recover the tokens with lower weight first, then recover the tokens that are strongly related to the classification task later.



The probability of a token being masked is tied to its attention score relative to the [CLS] token, reflecting its contribution to the TC objective. Figure \ref{probMask} shows that masking probabilities depend on the token's label-related information. Label-Aware Noise Scheduling guides the model to recover the most label-related key tokens from those less crucial to the classification task.

\subsubsection{Label-Aware Prompting}
\label{LAP}
However, such a noise schedule still poses a challenge to the conditional generation process. The diversity-consistency trade-off becomes more intense when important tokens are masked. With fewer unmasked tokens provided, the model naturally has a higher possibility of generating tokens that would break the label consistency.

\begin{figure}[h!]
    \centering
    \includegraphics[width=.5\textwidth]{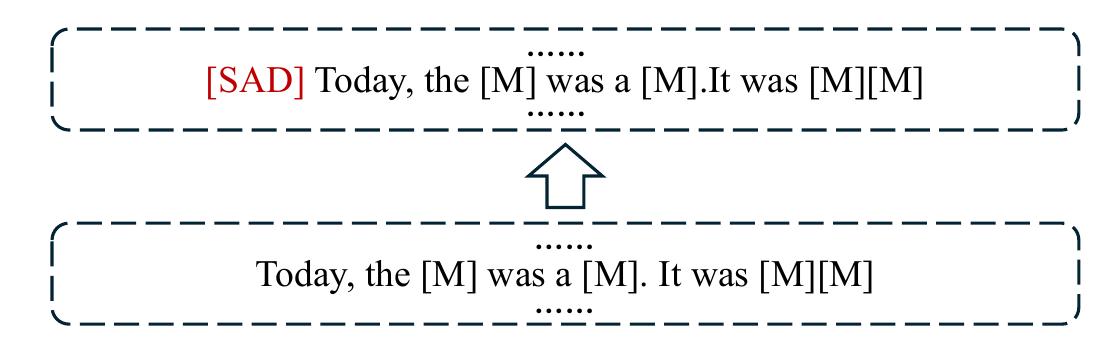}
    \caption{Label-Aware Prompting, each masked sequence is concatenated with their corresponding label.}
    \label{prompting}
        \vspace{-8pt}
\end{figure}

To address this challenge, we propose Label-Aware Prompting, a method that offers supplementary conditional information during both training and inference phrases. This additional information aids the model in generating samples that uphold label consistency.

As Figure \ref{prompting} illustrated, following the masking of samples in the noise schedule process, the labels of these samples are concatenated with their respective masked sequences. 

\subsection{Text Classification Model}


In this work, we adopt encoder-based PLM as our backbone model and finetuned them for the TC task. Though diffusion LM is strong enough to maintain consistency and diversity at the same time, the introduction of pseudo samples unavoidably introduced noise data to the training of the TC model. To mitigate such a problem, we design a contrastive learning-based noise-resistant training method, further improving the scalability of the proposed DiffusionCLS.

\subsubsection{Reflective Conditional Sample Generation}
We implement label prompting as a prior for the sample generator, akin to Label-Aware Prompting. Additionally, we introduce a novel reflective conditional sample generation module within the training loop of the TC model. This module dynamically generates masked sequences for the sample generator, integrating insights from label annotations and attention scores derived from the TC model simultaneously, calculating weights for each token with Eq.\ref{tokenWCal}.



However, generating pseudo samples from varying degrees of masking will result in various degrees of context replacement flexibility, thus impacting the consistency and diversity of pseudo samples. Essentially, providing a proper amount of conditional information will lead to plausible samples. Thus, we perform multiple experiments to search for the best condition, which will be further discussed in Section \ref{discussion}.

\subsubsection{Noise-Resistant Training}
\label{NRT}
The introduction of pseudo samples unavoidably introduced noise data to the training of the TC model. To mitigate such a problem, we design a contrastive learning-based noise-resistant training method, further improving the scalability of the proposed DiffusionCLS.

\begin{figure}[h!]
    \centering
    \includegraphics[width=.45\textwidth]{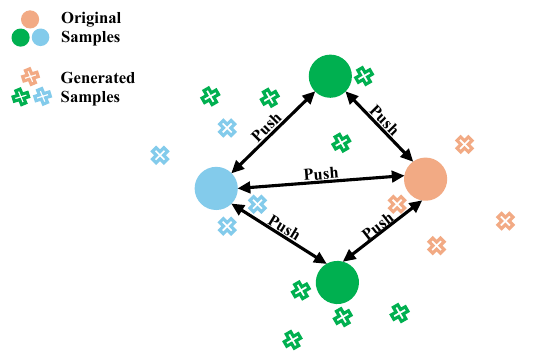}
    \caption{Noise-resistant contrastive learning. Cross points are generated samples while round dots denote original samples. Train-with-noise objective aiming at enlarging the gap between original samples with different labels.}
    \label{cl}
\end{figure}

Figure \ref{cl} demonstrates the Noise-resistant Training. Specifically, besides including supervision signals from labels of original and generated samples, we also guide the model to enlarge the gap between samples with different labels.

Consider a dataset comprising $m$ distinct categories $C = \{c_1,c_2,...,c_m\}$, we can obtain $k$ samples from the original training set, and the corresponding subscript list is $I = \{1, 2,..., k-1, k\}$. Essentially, a batch of sentences $S = \{s_1, s_2, ..., s_{k-1}, s_k\}$, their corresponding label sequence $L = [l_1, l_2, ..., l_{k-1}, l_{k}]$ with $l_i \in C$, and negative set for each sample $N_i = \{j \in I| l_j\neq l_i\}$.  From this, we derive semantic representations $H = \{h_1, h_2, ..., h_{k-1}, h_{k}\}$ from the TC model. Furthermore, employing a sample generator yields $B$ new samples for each original sample $s_i$, denoted as $G_i = \{g_0^{s_i},g_1^{s_i}, ..., g_{B-1}^{s_i}, g_B^{s_i}\}$, where $g_0^{s_i} = s_i$.

\textbf{Contrastive Loss.} To avoid expanding the impact of noise samples, we calculate contrastive loss from only the original samples. With the aim to enlarge the gap between samples from different categories, the contrastive loss can be calculated as:
\begin{equation}
    L_c = \frac{1}{K} \mathrm{log} \sum_{i\in I} \sum_{j \in N_{i}} \mathrm{exp}(\frac{\mathrm{sim}(h_i,h_j)}{\tau})  ,
\end{equation}
where $\mathrm{sim}()$ denotes the consine similarity function and $\tau$ is a  hyper-parameter as a scaling term.

\textbf{Classification Loss.} We also allows supervision signals directly affects the training of the TC model through the cross entropy loss, which can be denoted as:
    \vspace{-10pt}

\begin{equation}
    L_e = -\frac{1}{K(B+1)} \sum_{i\in I} \sum_{b=0}^{B} \sum_{c\in C} y^i_{b,c} \mathrm{log}(\hat{y}^i_{b,c}),
\end{equation}
where $y^i_{b,c}$ is the label indicator, and $\hat{y}^i_{b,c}$ is the predicted probability of $b$-th pseudo sample of the original sample $i$ being of class $c$.

\textbf{Training Objective.} From two losses mentioned above, we formulated the overall training objective for the TC model, which can be denoted as:

\begin{equation}
    L = L_c + L_e.
\end{equation}

\section{Experiments}
\label{experiments}

\subsection{Datasets and Baselines}

To measure the efficiency of the propose DiffusionCLS, we utilize both domain-specific and domain-general datasets comprising samples in Chinese, English, Arabic, French, and Spanish. Namely, domain-specific SMP2020-EWECT\footnote{https://smp2020ewect.github.io}, India-COVID-X\footnote{https://www.kaggle.com/datasets/surajkum1198/twitterdata}, SenWave \cite{yang2020senwave}, and domain-general SST-2 \cite{maas-EtAl:2011:ACL-HLT2011}. Additionally, to compare with the most cutting-edge low-resource TC methods, we utilize SST-2 dataset to evaluate our method in the few-shot setting. Dataset statistic and descriptions are demonstrated in Appendix \ref{app:datastatistics}.

To thoroughly explore and validate the capabilities of DiffusionCLS, we compare our method with a range of data augmentation techniques, from classic approaches to the latest advancements for low-resource TC. Specifically, we take Resample, Back Translation \cite{shleifer2019low}, Easy Data Augmentation (EDA) \cite{wei-zou-2019-eda}, SFT GPT-2 referenced to LAMBADA \cite{anabytavor2019data}, AEDA \cite{karimi-etal-2021-aeda-easier}, and GENIUS \cite{guo2022genius} as our baselines. Also, we compare our method in the few-shot setting with a couple of cutting-edge methods, namely, {SSMBA} \cite{ng-etal-2020-ssmba}, {ALP} \cite{kim2022alp}, and {SE} \cite{zheng-etal-2023-self}. More details of our baselines are demonstrated in Appendix \ref{apx:baselines}.

\begin{table*}[h!]
  \centering

\scalebox{0.83}{
    \begin{tabular}{c|c|cccc|cccc}
    \toprule
    \multirow{2}[4]{*}{Methods} & \multirow{2}[4]{*}{Policy} & \multicolumn{4}{c|}{SMP2020-EWECT} & \multicolumn{4}{c}{India-COVID-X} \\
\cmidrule{3-10}          &       & Macro-F & Acc   & $\Delta \mathrm{F}$   & $\Delta \mathrm{Acc}$ & Macro-F & Acc   & $\Delta \mathrm{F}$   & $\Delta \mathrm{Acc}$ \\
    \midrule
    Raw PLM   & N/A   & 65.87\% & 79.17\% & -     & -     & 70.99\% & 70.63\% & -     & - \\
    \midrule
    + Resample & B/D   & 64.84\% & 78.17\% & \textcolor[rgb]{ .753,  0,  0}{-1.03\%} & \textcolor[rgb]{ .753,  0,  0}{-1.00\%} & 72.74\% & 72.57\% & \textcolor[rgb]{ 0,  .69,  .314}{1.75\%} & \textcolor[rgb]{ 0,  .69,  .314}{1.94\%} \\
    
    + BT \citeyearpar{shleifer2019low} & B/D   & 64.03\% & 77.93\% & \textcolor[rgb]{ .753,  0,  0}{-1.84\%} & \textcolor[rgb]{ .753,  0,  0}{-1.24\%} & 72.93\% & 72.79\% & \textcolor[rgb]{ 0,  .69,  .314}{1.94\%} & \textcolor[rgb]{ 0,  .69,  .314}{2.16\%} \\
    
    + EDA \citeyearpar{wei-zou-2019-eda} & B/D   & 65.88\% & 78.87\% & \textcolor[rgb]{ 0,  .69,  .314}{0.01\%} & \textcolor[rgb]{ .753,  0,  0}{-0.30\%} & 66.83\% & 66.41\% & \textcolor[rgb]{ .753,  0,  0}{-4.16\%} & \textcolor[rgb]{ .753,  0,  0}{-4.22\%} \\
    
    + AEDA \citeyearpar{karimi-etal-2021-aeda-easier} & B/D   & \textit{66.58}\% & \textit{79.50}\% & \textcolor[rgb]{ 0,  .69,  .314}{\textit{0.71}\%} & \textcolor[rgb]{ 0,  .69,  .314}{\textit{0.33}\%} & {72.90\%} & {72.89\%} & \textcolor[rgb]{ 0,  .69,  .314}{1.91\%} & \textcolor[rgb]{ 0,  .69,  .314}{2.26\%} \\

    + GENIUS \citeyearpar{guo2022genius} & B/D   &    {64.27\%}   &   {78.23\% }   &   \textcolor[rgb]{ .753,  0,  0}{-1.60\%}    &    \textcolor[rgb]{ .753,  0,  0}{-0.94\%}   & 72.84\% & 72.46\% & \textcolor[rgb]{ 0,  .69,  .314}{1.85\%} & \textcolor[rgb]{ 0,  .69,  .314}{1.83\%} \\

    + DiffusionCLS (ours) & B/D   & {66.47\%} & {79.43\%} & \textcolor[rgb]{ 0,  .69,  .314}{0.60\%} & \textcolor[rgb]{ 0,  .69,  .314}{0.26\%} & 72.80\% & 72.57\% & \textcolor[rgb]{ 0,  .69,  .314}{{1.81\%}} & \textcolor[rgb]{ 0,  .69,  .314}{{1.94\%}} \\
    
    \midrule
    + BT \citeyearpar{shleifer2019low} & G/E   & 65.15\% & 77.93\% & \textcolor[rgb]{ .753,  0,  0}{-0.72\%} & \textcolor[rgb]{ .753,  0,  0}{-1.24\%} & 74.40\% & 74.30\% & \textcolor[rgb]{ 0,  .69,  .314}{3.41\%} & \textcolor[rgb]{ 0,  .69,  .314}{3.67\%} \\
    
    + EDA \citeyearpar{wei-zou-2019-eda} & G/E   & 50.12\% & 71.87\% & \textcolor[rgb]{ .753,  0,  0}{-15.75\%} & \textcolor[rgb]{ .753,  0,  0}{-7.30\%} & 74.15\% & 73.87\% & \textcolor[rgb]{ 0,  .69,  .314}{3.16\%} & \textcolor[rgb]{ 0,  .69,  .314}{3.24\%} \\
    
    + GPT-2 \citeyearpar{anabytavor2019data} & G/E   & 65.06\% & 77.80\% & \textcolor[rgb]{ .753,  0,  0}{-0.81\%} & \textcolor[rgb]{ .753,  0,  0}{-1.37\%} & 69.55\% & 69.58\% & \textcolor[rgb]{ .753,  0,  0}{-1.44\%} & \textcolor[rgb]{ .753,  0,  0}{-1.05\%} \\
    
    + AEDA \citeyearpar{karimi-etal-2021-aeda-easier} & G/E   & 65.81\% & 78.93\% & \textcolor[rgb]{ .753,  0,  0}{-0.06\%} & \textcolor[rgb]{ .753,  0,  0}{-0.24\%} & \textbf{75.49\%} & \textbf{75.27\%} & \textcolor[rgb]{ 0,  .69,  .314}{\textbf{4.50\%}} & \textcolor[rgb]{ 0,  .69,  .314}{\textbf{4.64\%}} \\
    
    + GENIUS \citeyearpar{guo2022genius} & G/E   &    {64.30\%}   &   {78.07\% }   &   \textcolor[rgb]{ .753,  0,  0}{-1.57\%}    &    \textcolor[rgb]{ .753,  0,  0}{-1.10\%}   & 74.28\% & 74.08\% & \textcolor[rgb]{ 0,  .69,  .314}{3.29\%} & \textcolor[rgb]{ 0,  .69,  .314}{3.45\%} \\
    
    + DiffusionCLS (ours) & G/E   & \textbf{67.98\%} & \textbf{80.23\%} & \textcolor[rgb]{ 0,  .69,  .314}{\textbf{2.11\%}} & \textcolor[rgb]{ 0,  .69,  .314}{\textbf{1.06\%}} & {\textit{74.65}\%} & {\textit{74.41}\%} & \textcolor[rgb]{ 0,  .69,  .314}{\textit{3.66}\%} & \textcolor[rgb]{ 0,  .69,  .314}{\textit{3.78}\%} \\
    
    \bottomrule
    \end{tabular}%

}
    \caption{Experiment results on SMP2020-EWECT and India-COVID-X datasets, with N/A indicating no augmentation, B/D for balancing pseudo samples, and G/E for the n-samples-each policy. We adopt bert-base as the English PLM and wwm-roberta as the Chinese PLM. + denotes the model is trained with the corresponding augmentation method. $\Delta$Acc and $\Delta$F represent performance variance between training with augmentation and without. }
  \label{BASELINES}%
\end{table*}%

\begin{table*}[ht]
  \centering
  \scalebox{0.78}{
  
\begin{tabular}{c|c|cccccccccc}
    \toprule
    \multirow{3}[4]{*}{Dataset} & \multirow{3}[4]{*}{\#Shot} & \multicolumn{10}{c}{Data Augmentaion Methods} \\
\cmidrule{3-12}          &       & N/A   & \makecell{+EDA${}^\dagger$\\ \citeyearpar{wei-zou-2019-eda}}  & \makecell{+BT${}^\dagger$\\ \citeyearpar{shleifer2019low}}   & \makecell{+SSMBA${}^\dagger$\\ \citeyearpar{ng-etal-2020-ssmba}} & \makecell{+ALP${}^\dagger$\\ \citeyearpar{kim2022alp}}  & \makecell{+SE${}^\dagger$\\ \citeyearpar{zheng-etal-2023-self}} & \makecell{+GPT-2\\ \citeyearpar{anabytavor2019data}} & \makecell{+mixup\\ \citeyearpar{zhang2017mixup}}& \makecell{+AWD\\ \citeyearpar{chen2023adversarial}} & \makecell{+DiffusionCLS} \\
    \midrule
    \multirow{2}[4]{*}{SST-2} & 5     & 54.38 & 56.22 & 55.77 & 56.34 & \textit{63.40}  & -    & 52.18 & 61.81 & 58.86 & \textbf{65.30} \\
    \cmidrule{2-12}          & 10    & 61.82 & 53.96 & 62.05 & 59.05 & \textbf{69.72} & 57.56 & 54.17 & 61.55 & 64.62 & \textit{68.29} \\
    \bottomrule
    \end{tabular}%
  
  }
    
      \caption{Performances of TC models on dataset SST-2 under the few-shot setting. $\dagger$ denotes that results are from previous research. All of our results are collected from 5 runs with different seeds.}
  \label{few-shot}%
\end{table*}%

\subsection{Experiment Setup}
We set up two experimental modes, entire data mode and partial data mode, to reveal the effectiveness of our method in different scenarios. Since severe imbalanced distribution challenges existed, we take macro-F1 and accuracy as our major evaluation metrics.

Also, we conduct 5-shot and 10-shot experiments on SST-2 to investigate the performance of DiffusionCLS in extreme low-resource conditions. For evaluation, we use accuracy as the metric and report the average results over three random seeds to minimize the effects of stochasticity.

Additionally, we setup comparisons between variant augmentation policies, namely, generate new samples until the dataset distribution is balanced, and generate n pseudo samples for each sample (n-samples-each), which denoted as B/D and G/E in Table \ref{BASELINES}, and n=4 in our experiments.
Other related implementation details are described in Appendix \ref{app:datastatistics}.

\subsection{Results and Analysis}
\begin{figure*}[h!]
    \centering
    \begin{subfigure}{0.25\textwidth}
        \centering
        \includegraphics[width=\textwidth]{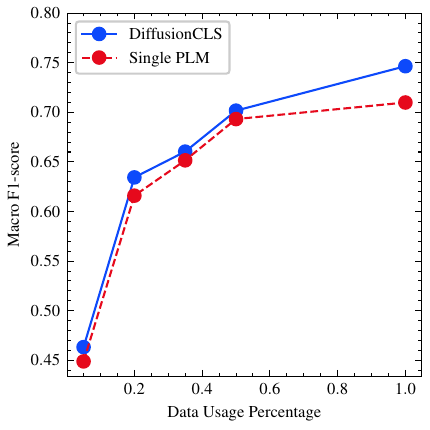}
        \caption{Arabic}
    \end{subfigure}
    \hspace{1em}
    \begin{subfigure}{0.25\textwidth}
        \centering
        \includegraphics[width=\textwidth]{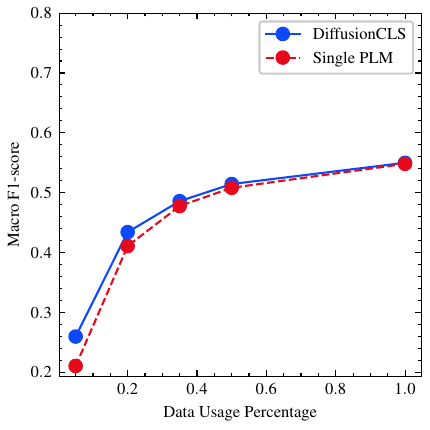}
        \caption{Spanish}
    \end{subfigure}
    \hspace{1em}
    \begin{subfigure}{0.25\textwidth}
        \centering
        \includegraphics[width=\textwidth]{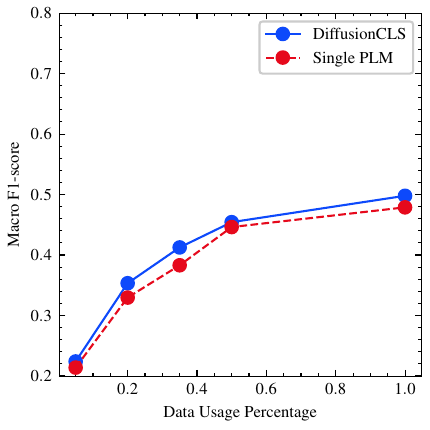}
        \caption{French}
    \end{subfigure}
    \caption{Performances of SC models on dataset SenWave under the partial data setting. Red lines denote the raw PLM results and blue lines represent models trained with DiffusionCLS.}
    \label{senwave}
\end{figure*}

The results of entire-data-setting experiments on datasets SMP2020-EWECT and India-COVID-X are mainly demonstrated in Table \ref{BASELINES}, which we compare DiffusionCLS with other strong DA baselines. For experiments with partial-data and few-shot settings, results are majorly showed in Figure \ref{senwave} and Table \ref{low-resource-cn-eng-appendix}.

\textbf{Results under Entire Data Mode.} As shown in Table \ref{BASELINES}, in general, the proposed DiffusionCLS outperforms most DA methods on domain-specific datasets SMP2020-EWECT and India-COVID-X, especially under G/E policy. Notably, the DiffusionCLS positively impacts the TC model across all policies and datasets, which most baselines fail.

Our method excels in dealing with the challenge of uneven datasets. Under severe uneven distribution and domain-specific scenarios, i.e., the dataset SMP2020-EWECT, most DA baselines fail to impact the classification model positively except DiffusionCLS, which achieves the best performance. Also, our method achieves competitive performance under data-sparse and domain-specific scenarios, i.e., in the dataset India-COVID-X, most DA methods bring improvement to the classification model, and our DiffusionCLS ranked second.

Rule-based DA methods such as EDA, rather lack diversity bringing overfit problems or solely relying on out-domain knowledge therefore breaking consistency and impacting the task model negatively. For model-based methods, though most methods significantly increase the diversity of the generated samples, they rather generate samples solely depending on pretraining knowledge and in-context-learning techniques or generate samples only conditioned on the label itself, posing a challenge of maintaining consistency. 


\textbf{Results under Partial Data Mode and Few-shot Settings.} As shown in Figure \ref{senwave} and Table \ref{low-resource-cn-eng-appendix} in Appendix \ref{low-r-table}, the proposed DiffusionCLS method consistently improves the classification model. Notably, DiffusionCLS matches the PLM baseline performance on the Arabic SenWave dataset using only 50\% of the data samples.

We also compare DiffusionCLS with the most cutting-edge few-shot methods on SST-2 dataset under 5-shot and 10-shot setting, the results are shown in Table \ref{few-shot}. Though our method fails to surpass all few-shot baselines, it still achieves competitive performance with those designed for the few-shot task. 

Since DiffusionCLS requires diffusion training to adapt to domain-specific tasks, extreme sample insufficiency may introduce noise, negatively impacting the model. However, our method positively impacts the TC model in most low-resource cases by effectively utilizing pre-trained and in-domain knowledge, from severe imbalanced label distribution to severe sample insufficiency.

\subsection{Ablation Study}

To validate the effectiveness of modules in the proposed DiffusionCLS, we conduct ablation studies to study the impacts of each module. Table \ref{ablation_result} presents the results of the ablation experiments. In each row of the experiment results, one of the modules in DiffusionCLS has been removed for discussion, except D.A., which removes all modules related to the generator and only applies noise-resistance training.

Overall, all modules in the proposed DiffusionCLS works positively to the TC model, compared with the pure PLM model, the application of DiffusionCLS leads to 2.11\% and 3.66\% rises in F1 values on dataset SMP2020-EWECT and India-COVID-X respectively.

The results of ablation studies further validate that the Label-Aware Prompting effectively improves the quality of pseudo samples. Also, the Noise-Resistant Training reduces the impact of noise pseudo samples.

\begin{table}[h!]
  \centering
  \scalebox{0.9}{
    \begin{tabular}{c|c|c|c}
    \toprule
    Dataset & Methods & Macro-F & Acc \\
    \midrule
    \multirow{4}[2]{*}{\makecell{SMP2020-\\EWECT}} 
          & DiffusionCLS  & \textbf{0.6798 } & \textbf{0.8023 } \\
          & -w/o D.A. & 0.6637  & 0.7957  \\
          & -w/o L.A.P. & 0.6671  & 0.7930  \\
          & -w/o N.R.T. & 0.6695  & 0.7963  \\
    \midrule
    \multirow{4}[2]{*}{\makecell{India-\\COVID-X}} 
          & DiffusionCLS  & \textbf{0.7465 } & \textbf{0.7441 } \\
          & -w/o D.A. & 0.7206  & 0.7181  \\
          & -w/o L.A.P. & 0.7298  & 0.7268  \\
          & -w/o N.R.T. & 0.7361  & 0.7354  \\
    \bottomrule
    \end{tabular}%
  }
      \caption{Experiment results of ablation study, where -w/o is the abbreviation of without. D.A., \hyperref[LAP]{L.A.P.}, and \hyperref[NRT]{N.R.T.} correspond to data augmentation, label-aware prompting, and noise-resistant training. D.A. removes the generator.}
  \label{ablation_result}%
\end{table}%

\subsection{Discussions and Visualizations}
\label{discussion}
Generating pseudo samples from more masked tokens provides more flexibility for generation and tends to result in more diverse samples, however, it will enlarge the possibility of breaking the consistency since less information is provided.

To analyze the optimal amount of masks for generating new pseudo samples, we conduct experiments on the India-COVID-X dataset. During conditional sample generation, we gather masked sequences from 32 noise-adding steps, group them into sets of eight, and evaluate how varying masking levels impact the model's performance.

\begin{figure}[h!]
    \centering
    \includegraphics[width=.45\textwidth]{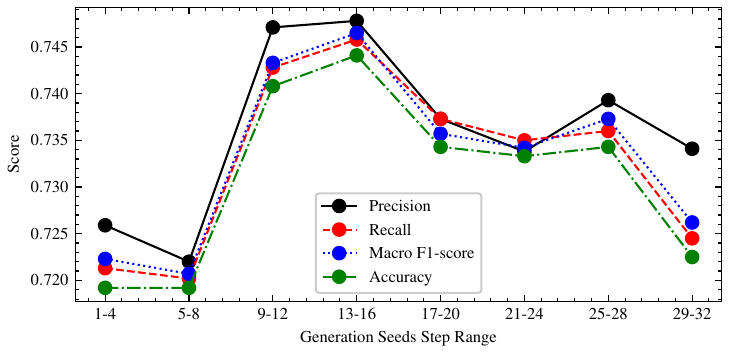}
    \caption{Performances of models with pseudo samples generated from different groups of masked sequences, in which step one will result in original sequences and step 32 will result in generating pseudo samples from fully masked sequences.}
    \label{perOverSteps}
    \vspace{-8pt}
\end{figure}

As shown in Figure \ref{perOverSteps}, our observations indicate a unimodal trend. The model's performance improves with increased masking, peaks at the 4th group, and then declines with further masking. This reflects the diversity-consistency trade-off, more masked tokens create more diverse samples, but overly diverse samples may be inconsistent with original labels or domain.

\begin{figure}[h!]
    \centering
    \includegraphics[width=.45\textwidth]{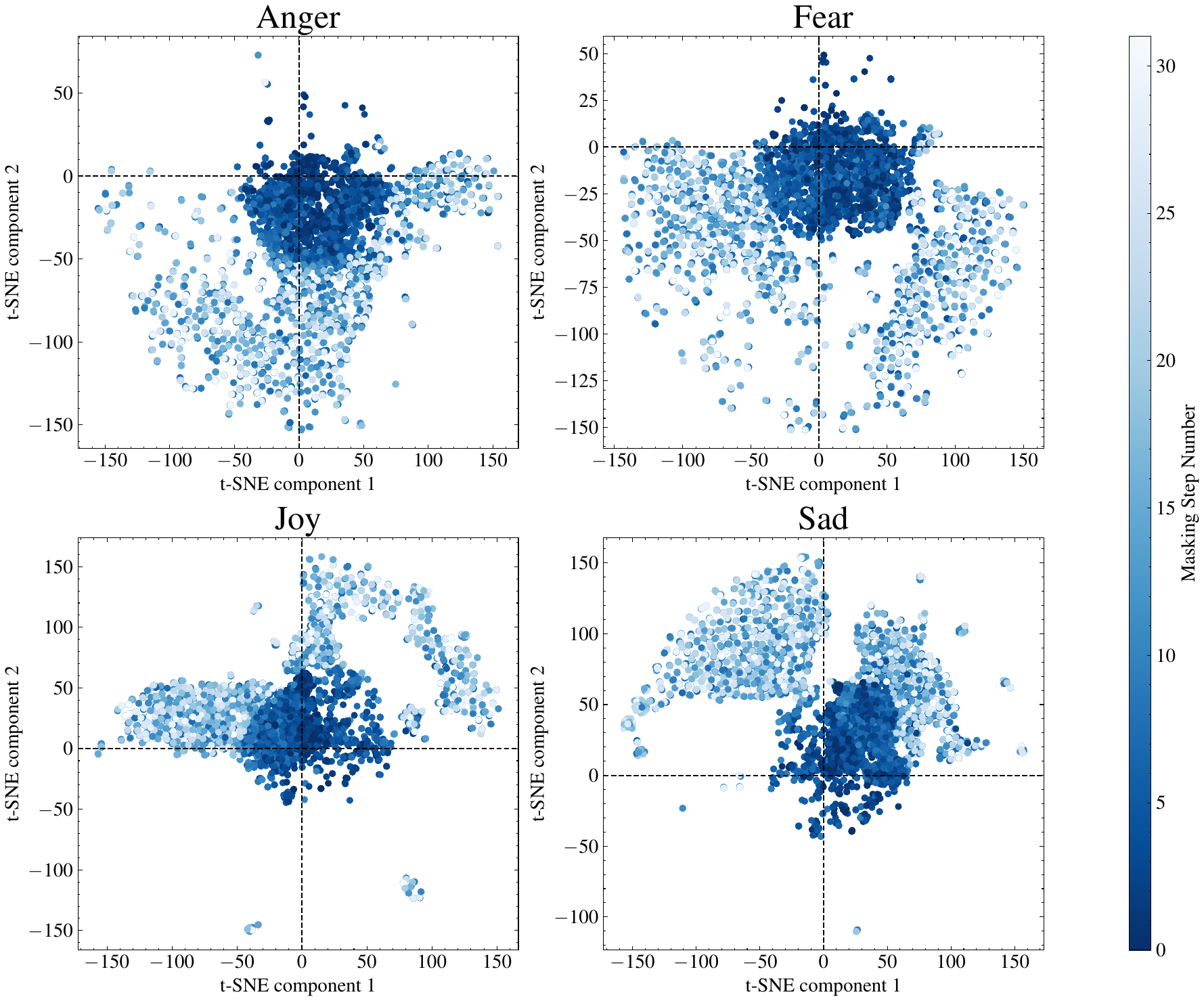}
    \caption{2D t-SNE visualization on the India-COVID-X dataset.}
    \label{t-SNE}
\end{figure}

To explore the relationship between generated pseudo samples and original samples, we conduct 2D t-SNE visualization. Figure \ref{t-SNE} shows that as masking increases, pseudo samples gradually diverge from the original samples, indicating increased diversity. 

\section{Conclusion}
In this work, we propose DiffusionCLS, a novel approach tackling SC challenges under low-resource conditions, especially in domain-specific and uneven distribution scenarios. Utilizing a diffusion LM, DiffusionCLS captures in-domain knowledge to generate high-quality pseudo samples maintaining both diversity and consistency. This method surpasses various kinds of data augmentation techniques. Our experiments demonstrate that DiffusionCLS significantly enhances SC performance across various domain-specific and multilingual datasets. Ablation and visualization studies further validate our approach, emphasizing the importance of balancing diversity and consistency in pseudo samples. DiffusionCLS presents a robust solution for data augmentation in low-resource NLP applications, paving a promising path for future research.

\section*{Limitations}
Like most model-based data augmentation methods, the performance of data generators is also limited in extreme low-resource scenarios. This limitation persists because the model still necessitates training on the training data, even with the potential expansion of the dataset through the inclusion of unlabeled data, data deficiency impacts the data generator negatively.

\section*{Acknowledgments}
This work was supported by the National Social Science Fund of China (No. 22BTQ045).

\bibliography{acl_latex}

\begin{thebibliography}{35}
\providecommand{\natexlab}[1]{#1}

\bibitem[{Anaby-Tavor et~al.(2020)Anaby-Tavor, Carmeli, Goldbraich, Kantor, Kour, Shlomov, Tepper, and Zwerdling}]{anabytavor2019data}
Ateret Anaby-Tavor, Boaz Carmeli, Esther Goldbraich, Amir Kantor, George Kour, Segev Shlomov, Naama Tepper, and Naama Zwerdling. 2020.
\newblock Do not have enough data? deep learning to the rescue!
\newblock In \emph{Proceedings of the AAAI conference on artificial intelligence}, volume~34, pages 7383--7390.

\bibitem[{Chen et~al.(2023{\natexlab{a}})Chen, Zhang, Luo, Hu, and Mao}]{chen2023adversarial}
Junfan Chen, Richong Zhang, Zheyan Luo, Chunming Hu, and Yongyi Mao. 2023{\natexlab{a}}.
\newblock Adversarial word dilution as text data augmentation in low-resource regime.
\newblock In \emph{Proceedings of the AAAI Conference on Artificial Intelligence}, volume~37, pages 12626--12634.

\bibitem[{Chen et~al.(2018)Chen, Sun, Athiwaratkun, Cardie, and Weinberger}]{chen-etal-2018-adversarial}
Xilun Chen, Yu~Sun, Ben Athiwaratkun, Claire Cardie, and Kilian Weinberger. 2018.
\newblock \href {https://doi.org/10.1162/tacl_a_00039} {Adversarial deep averaging networks for cross-lingual sentiment classification}.
\newblock \emph{Transactions of the Association for Computational Linguistics}, 6:557--570.

\bibitem[{Chen et~al.(2023{\natexlab{b}})Chen, Tian, Wang, and Jiang}]{chen2023distantly}
Zhuowei Chen, Yujia Tian, Lianxi Wang, and Shengyi Jiang. 2023{\natexlab{b}}.
\newblock A distantly-supervised relation extraction method based on selective gate and noise correction.
\newblock In \emph{China National Conference on Chinese Computational Linguistics}, pages 159--174. Springer.

\bibitem[{Devlin et~al.(2018)Devlin, Chang, Lee, and Toutanova}]{devlin2018bert}
Jacob Devlin, Ming-Wei Chang, Kenton Lee, and Kristina Toutanova. 2018.
\newblock Bert: Pre-training of deep bidirectional transformers for language understanding.
\newblock \emph{arXiv preprint arXiv:1810.04805}.

\bibitem[{Feng et~al.(2021)Feng, Gangal, Wei, Chandar, Vosoughi, Mitamura, and Hovy}]{feng2021survey}
Steven~Y Feng, Varun Gangal, Jason Wei, Sarath Chandar, Soroush Vosoughi, Teruko Mitamura, and Eduard Hovy. 2021.
\newblock A survey of data augmentation approaches for nlp.
\newblock In \emph{Findings of the Association for Computational Linguistics: ACL-IJCNLP 2021}, pages 968--988.

\bibitem[{Gatto and Preum(2023)}]{gatto2023not}
Joseph Gatto and Sarah~M Preum. 2023.
\newblock Not enough labeled data? just add semantics: A data-efficient method for inferring online health texts.
\newblock \emph{arXiv preprint arXiv:2309.09877}.

\bibitem[{Goudjil et~al.(2018)Goudjil, Koudil, Bedda, and Ghoggali}]{goudjil2018novel}
Mohamed Goudjil, Mouloud Koudil, Mouldi Bedda, and Noureddine Ghoggali. 2018.
\newblock A novel active learning method using svm for text classification.
\newblock \emph{International Journal of Automation and Computing}, 15:290--298.

\bibitem[{Guo et~al.(2022)Guo, Gong, Shen, Han, Huang, Duan, and Chen}]{guo2022genius}
Biyang Guo, Yeyun Gong, Yelong Shen, Songqiao Han, Hailiang Huang, Nan Duan, and Weizhu Chen. 2022.
\newblock Genius: Sketch-based language model pre-training via extreme and selective masking for text generation and augmentation.
\newblock \emph{arXiv preprint arXiv:2211.10330}.

\bibitem[{He et~al.(2023)He, Sun, Tang, Wang, Huang, and Qiu}]{he-etal-2023-diffusionbert}
Zhengfu He, Tianxiang Sun, Qiong Tang, Kuanning Wang, Xuanjing Huang, and Xipeng Qiu. 2023.
\newblock \href {https://doi.org/10.18653/v1/2023.acl-long.248} {{D}iffusion{BERT}: Improving generative masked language models with diffusion models}.
\newblock In \emph{Proceedings of the 61st Annual Meeting of the Association for Computational Linguistics (Volume 1: Long Papers)}, pages 4521--4534, Toronto, Canada. Association for Computational Linguistics.

\bibitem[{Ho et~al.(2020)Ho, Jain, and Abbeel}]{ho2020denoising}
Jonathan Ho, Ajay Jain, and Pieter Abbeel. 2020.
\newblock Denoising diffusion probabilistic models.
\newblock \emph{Advances in neural information processing systems}, 33:6840--6851.

\bibitem[{Karimi et~al.(2021)Karimi, Rossi, and Prati}]{karimi-etal-2021-aeda-easier}
Akbar Karimi, Leonardo Rossi, and Andrea Prati. 2021.
\newblock \href {https://doi.org/10.18653/v1/2021.findings-emnlp.234} {{AEDA}: An easier data augmentation technique for text classification}.
\newblock In \emph{Findings of the Association for Computational Linguistics: EMNLP 2021}, pages 2748--2754, Punta Cana, Dominican Republic. Association for Computational Linguistics.

\bibitem[{Kim et~al.(2022)Kim, Woo, Oh, Cha, and Han}]{kim2022alp}
Hazel~H Kim, Daecheol Woo, Seong~Joon Oh, Jeong-Won Cha, and Yo-Sub Han. 2022.
\newblock Alp: Data augmentation using lexicalized pcfgs for few-shot text classification.
\newblock In \emph{Proceedings of the aaai conference on artificial intelligence}, volume~36, pages 10894--10902.

\bibitem[{Kumar et~al.(2020)Kumar, Choudhary, and Cho}]{kumar2020data}
Varun Kumar, Ashutosh Choudhary, and Eunah Cho. 2020.
\newblock Data augmentation using pre-trained transformer models.
\newblock In \emph{Proceedings of the 2nd Workshop on Life-long Learning for Spoken Language Systems}, pages 18--26.

\bibitem[{Lakshmi and Velmurugan(2023)}]{lakshmi2023classification}
S~Deepa Lakshmi and T~Velmurugan. 2023.
\newblock Classification of disaster tweets using natural language processing pipeline.
\newblock \emph{Acta Scientific COMPUTER SCIENCES Volume}, 5(3).

\bibitem[{Lan et~al.(2019)Lan, Chen, Goodman, Gimpel, Sharma, and Soricut}]{lan2019albert}
Zhenzhong Lan, Mingda Chen, Sebastian Goodman, Kevin Gimpel, Piyush Sharma, and Radu Soricut. 2019.
\newblock Albert: A lite bert for self-supervised learning of language representations.
\newblock \emph{arXiv preprint arXiv:1909.11942}.

\bibitem[{Liu et~al.(2019)Liu, Ott, Goyal, Du, Joshi, Chen, Levy, Lewis, Zettlemoyer, and Stoyanov}]{liu2019roberta}
Yinhan Liu, Myle Ott, Naman Goyal, Jingfei Du, Mandar Joshi, Danqi Chen, Omer Levy, Mike Lewis, Luke Zettlemoyer, and Veselin Stoyanov. 2019.
\newblock Roberta: A robustly optimized bert pretraining approach.
\newblock \emph{arXiv preprint arXiv:1907.11692}.

\bibitem[{Maas et~al.(2011)Maas, Daly, Pham, Huang, Ng, and Potts}]{maas-EtAl:2011:ACL-HLT2011}
Andrew~L. Maas, Raymond~E. Daly, Peter~T. Pham, Dan Huang, Andrew~Y. Ng, and Christopher Potts. 2011.
\newblock \href {http://www.aclweb.org/anthology/P11-1015} {Learning word vectors for sentiment analysis}.
\newblock In \emph{Proceedings of the 49th Annual Meeting of the Association for Computational Linguistics: Human Language Technologies}, pages 142--150, Portland, Oregon, USA. Association for Computational Linguistics.

\bibitem[{Nabil et~al.(2023)Nabil, Das, Salim, Arifeen, and Fattah}]{nabil2023bangla}
Alvi~Ahmmed Nabil, Dola Das, Md~Shahidul Salim, Shamsul Arifeen, and HM~Abdul Fattah. 2023.
\newblock Bangla emergency post classification on social media using transformer based bert models.
\newblock In \emph{2023 6th International Conference on Electrical Information and Communication Technology (EICT)}, pages 1--6. IEEE.

\bibitem[{Ng et~al.(2020)Ng, Cho, and Ghassemi}]{ng-etal-2020-ssmba}
Nathan Ng, Kyunghyun Cho, and Marzyeh Ghassemi. 2020.
\newblock \href {https://doi.org/10.18653/v1/2020.emnlp-main.97} {{SSMBA}: Self-supervised manifold based data augmentation for improving out-of-domain robustness}.
\newblock In \emph{Proceedings of the 2020 Conference on Empirical Methods in Natural Language Processing (EMNLP)}, pages 1268--1283, Online. Association for Computational Linguistics.

\bibitem[{Nichol and Dhariwal(2021)}]{nichol2021improved}
Alexander~Quinn Nichol and Prafulla Dhariwal. 2021.
\newblock Improved denoising diffusion probabilistic models.
\newblock In \emph{International conference on machine learning}, pages 8162--8171. PMLR.

\bibitem[{Ogueji et~al.(2021)Ogueji, Zhu, and Lin}]{ogueji2021small}
Kelechi Ogueji, Yuxin Zhu, and Jimmy Lin. 2021.
\newblock Small data? no problem! exploring the viability of pretrained multilingual language models for low-resourced languages.
\newblock In \emph{Proceedings of the 1st Workshop on Multilingual Representation Learning}, pages 116--126.

\bibitem[{Patwa et~al.(2024)Patwa, Filice, Chen, Castellucci, Rokhlenko, and Malmasi}]{patwa-etal-2024-enhancing-low}
Parth Patwa, Simone Filice, Zhiyu Chen, Giuseppe Castellucci, Oleg Rokhlenko, and Shervin Malmasi. 2024.
\newblock \href {https://aclanthology.org/2024.lrec-main.533} {Enhancing low-resource {LLM}s classification with {PEFT} and synthetic data}.
\newblock In \emph{Proceedings of the 2024 Joint International Conference on Computational Linguistics, Language Resources and Evaluation (LREC-COLING 2024)}, pages 6017--6023, Torino, Italia. ELRA and ICCL.

\bibitem[{Sedinkina et~al.(2022)Sedinkina, Schmitt, and Sch{\"u}tze}]{sedinkina2022domain}
Marina Sedinkina, Martin Schmitt, and Hinrich Sch{\"u}tze. 2022.
\newblock Domain adaptation for sparse-data settings: What do we gain by not using bert?
\newblock \emph{arXiv e-prints}, pages arXiv--2203.

\bibitem[{Shleifer(2019)}]{shleifer2019low}
Sam Shleifer. 2019.
\newblock Low resource text classification with ulmfit and backtranslation.
\newblock \emph{arXiv preprint arXiv:1903.09244}.

\bibitem[{Sun et~al.(2020)Sun, Wang, Li, Feng, Tian, Wu, and Wang}]{sun2020ernie}
Yu~Sun, Shuohuan Wang, Yukun Li, Shikun Feng, Hao Tian, Hua Wu, and Haifeng Wang. 2020.
\newblock Ernie 2.0: A continual pre-training framework for language understanding.
\newblock In \emph{Proceedings of the AAAI conference on artificial intelligence}, volume~34, pages 8968--8975.

\bibitem[{Tan et~al.(2019)Tan, Yu, Wang, Wang, Potdar, Chang, and Yu}]{tan2019out}
Ming Tan, Yang Yu, Haoyu Wang, Dakuo Wang, Saloni Potdar, Shiyu Chang, and Mo~Yu. 2019.
\newblock Out-of-domain detection for low-resource text classification tasks.
\newblock In \emph{Proceedings of the 2019 Conference on Empirical Methods in Natural Language Processing and the 9th International Joint Conference on Natural Language Processing (EMNLP-IJCNLP)}, pages 3566--3572.

\bibitem[{Wang et~al.(2024)Wang, Tian, and Chen}]{wang-etal-2024-enhancing-hindi}
Lianxi Wang, Yujia Tian, and Zhuowei Chen. 2024.
\newblock \href {https://aclanthology.org/2024.lrec-main.528} {Enhancing {H}indi feature representation through fusion of dual-script word embeddings}.
\newblock In \emph{Proceedings of the 2024 Joint International Conference on Computational Linguistics, Language Resources and Evaluation (LREC-COLING 2024)}, pages 5966--5976, Torino, Italia. ELRA and ICCL.

\bibitem[{Wei and Zou(2019)}]{wei-zou-2019-eda}
Jason Wei and Kai Zou. 2019.
\newblock \href {https://doi.org/10.18653/v1/D19-1670} {{EDA}: Easy data augmentation techniques for boosting performance on text classification tasks}.
\newblock In \emph{Proceedings of the 2019 Conference on Empirical Methods in Natural Language Processing and the 9th International Joint Conference on Natural Language Processing (EMNLP-IJCNLP)}, pages 6382--6388, Hong Kong, China. Association for Computational Linguistics.

\bibitem[{Wen and Fang(2023)}]{wen2023augmenting}
Zhihao Wen and Yuan Fang. 2023.
\newblock Augmenting low-resource text classification with graph-grounded pre-training and prompting.
\newblock In \emph{Proceedings of the 46th International ACM SIGIR Conference on Research and Development in Information Retrieval}, pages 506--516.

\bibitem[{Wu et~al.(2019)Wu, Lv, Zang, Han, and Hu}]{wu2019conditional}
Xing Wu, Shangwen Lv, Liangjun Zang, Jizhong Han, and Songlin Hu. 2019.
\newblock Conditional bert contextual augmentation.
\newblock In \emph{Computational Science--ICCS 2019: 19th International Conference, Faro, Portugal, June 12--14, 2019, Proceedings, Part IV 19}, pages 84--95. Springer.

\bibitem[{Yang et~al.(2020)Yang, Alamro, Albaradei, Salhi, Lv, Ma, Alshehri, Jaber, Tifratene, Wang, Gojobori, Duarte, Gao, and Zhang}]{yang2020senwave}
Qiang Yang, Hind Alamro, Somayah Albaradei, Adil Salhi, Xiaoting Lv, Changsheng Ma, Manal Alshehri, Inji Jaber, Faroug Tifratene, Wei Wang, Takashi Gojobori, Carlos~M. Duarte, Xin Gao, and Xiangliang Zhang. 2020.
\newblock \href {https://arxiv.org/abs/2006.10842} {Senwave: Monitoring the global sentiments under the covid-19 pandemic}.
\newblock \emph{Preprint}, arXiv:2006.10842.

\bibitem[{Yu et~al.(2023)Yu, Zhao, and Xia}]{yu-etal-2023-cross}
Jianfei Yu, Qiankun Zhao, and Rui Xia. 2023.
\newblock \href {https://doi.org/10.18653/v1/2023.acl-long.81} {Cross-domain data augmentation with domain-adaptive language modeling for aspect-based sentiment analysis}.
\newblock In \emph{Proceedings of the 61st Annual Meeting of the Association for Computational Linguistics (Volume 1: Long Papers)}, pages 1456--1470, Toronto, Canada. Association for Computational Linguistics.

\bibitem[{Zhang et~al.(2017)Zhang, Cisse, Dauphin, and Lopez-Paz}]{zhang2017mixup}
Hongyi Zhang, Moustapha Cisse, Yann~N Dauphin, and David Lopez-Paz. 2017.
\newblock mixup: Beyond empirical risk minimization.
\newblock \emph{arXiv preprint arXiv:1710.09412}.

\bibitem[{Zheng et~al.(2023)Zheng, Zhong, Ding, Tian, Niu, Wang, Li, and Tao}]{zheng-etal-2023-self}
Haoqi Zheng, Qihuang Zhong, Liang Ding, Zhiliang Tian, Xin Niu, Changjian Wang, Dongsheng Li, and Dacheng Tao. 2023.
\newblock \href {https://doi.org/10.18653/v1/2023.emnlp-main.555} {Self-evolution learning for mixup: Enhance data augmentation on few-shot text classification tasks}.
\newblock In \emph{Proceedings of the 2023 Conference on Empirical Methods in Natural Language Processing}, pages 8964--8974, Singapore. Association for Computational Linguistics.

\end{thebibliography}

\appendix

\section{Experiment Setup, Implementation, and Dataset Statistics}\label{app:datastatistics}

\subsection{Experiment Setup}
The low-resource challenge in TC includes problems like insufficient annotated samples, domain-specific adaptation problems, and imbalanced distribution. To measure the capability of the proposed DiffusionCLS to mitigate these problems, we conduct experiments on three domain-specific datasets with respect to the problems mentioned above, as shown in Table \ref{low-resource-challenge}. 

\subsection{Implementation}
For implementation, we take bert-base-uncased\footnote{https://huggingface.co/google-bert/bert-base-uncased} and chinese-roberta-wwm\footnote{https://huggingface.co/hfl/chinese-roberta-wwm-ext} from the huggingface platform respectively for English and Chinese dataset training. Also, hyper-parameters settings of our work are demonstrated in Table \ref{tab:hpm} and Table \ref{tab:hpm_ds}.

\subsection{Datasets}
For our experiments, we utilize multilingual datasets, both domain-specific and domain-general, to evaluate the proposed DiffusionCLS. Data statistics and their challenges are demonstrated in Table \ref{low-resource-challenge} and Table \ref{datastat}.
\begin{itemize}
    \item \textbf{SMP2020-EWECT\footnote{https://smp2020ewect.github.io}.} This Chinese dataset includes 8,606 pandemic-related posts, categorized into neutral, happy, angry, sad, fear, and surprise, with highly imbalanced label distribution.
    \item \textbf{India-COVID-X\footnote{https://www.kaggle.com/datasets/surajkum1198/twitterdata}.} This dataset contains cleaned English tweets from India X platform on topics such as coronavirus, COVID-19, and lockdown. The tweets have been labeled into four sentiment categories with relatively balanced label distribution.
    \item \textbf{SenWave}\cite{yang2020senwave}. This dataset includes about 5,000 English tweets and approximately 3,000 Arabic tweets in the specific domain of the pandemic and lockdown, which are annotated with sentiment labels. English-translated French and Spanish annotated samples are also included. We extract all single label samples for experiments.
    \item \textbf{SST-2}\cite{maas-EtAl:2011:ACL-HLT2011}. It includes 11,855 movie review sentences parsed by the Stanford parser, with 215,154 unique phrases annotated by three human judges.
\end{itemize}

\section{Baselines}
\label{apx:baselines}

\begin{itemize}

    \item {Non-Generative Methods} 
    \begin{itemize}
        \item \textbf{SSMBA}\cite{ng-etal-2020-ssmba}. Uses a corruption and reconstruction function to augment data by filling in masked portions.
    
        \item \textbf{ALP}\cite{kim2022alp}. Employs Lexicalized Probabilistic Context-Free Grammars to generate syntactically diverse augmented samples.
        
        \item \textbf{SE}\cite{zheng-etal-2023-self}. Utilizes a self-evolution learning-based mixup technique to create adaptive pseudo samples for training.
        
        \item \textbf{AEDA}\cite{karimi-etal-2021-aeda-easier}. Randomly insert punctuations into the original sentences to produce new samples.
    \end{itemize}

    \item {Generative Methods}
    \begin{itemize}
            \item \textbf{GPT-2}\cite{anabytavor2019data}. Fine-tunes GPT-2 with prompt-based SFT, prompting labels to generate pseudo samples.

            \item \textbf{GENIUS}\cite{guo2022genius}. 
        A conditional text generation model using sketches as input, which can fill in the missing context for a given sketch.
        
    \end{itemize}

    \item {Representation Augmentation Methods}
    \begin{itemize}
        \item \textbf{mixup}\cite{zhang2017mixup}. Mixup is a representational DA technique that creates new training samples by linearly interpolating between pairs of examples and their labels.

        \item \textbf{AWD}\cite{chen2023adversarial}. AWD generates challenging positive examples for low-resource text classification by diluting strong positive word embeddings with unknown-word embeddings.
    \end{itemize}

\end{itemize}

\section{Experiment Results with Partial Data Mode}
\label{low-r-table}
The proposed DiffusionCLS method consistently enhances the classification model, achieving higher accuracy with only 50\% training data than the raw PLM on dataset SMP2020-EWECT. Detailed results are shown in Table \ref{low-resource-cn-eng-appendix}.

\begin{table*}[h!]
  \centering
    \begin{tabular}{cc}
    \toprule
    Parameter & Value \\
    \midrule
    Epoch of Proxy Model & 15 \\
    \# Diffusion Steps & 32 \\
    Index of Diffusion Group & 4 \\
    \# Aug. Samples & 4 \\
    Learning Rate    & 4e-06 \\
    Weight Decay & 0.01 \\
    \bottomrule
    \end{tabular}%
    \caption{Settings of hyperparameters, all values are identical across all datasets.}
        \label{tab:hpm}%
\end{table*}%

\begin{table*}[h!]
  \centering

  \scalebox{0.9}{
\begin{tabular}{cccc}
    \toprule
    Dataset & PLM Name & Epoch of DiffusionLM & Batch Size\\
    \midrule
    SMP2020-EWECT & chinese-roberta-wwm-ext & 1 & 60\\
    India-COVID-X & bert-base-uncased & 1 & 40\\
    SenWave-Arabic & CAMeL-Lab/bert-base-arabic-camelbert-ca & 2 & 60\\
    SenWave-France & dbmdz/bert-base-french-europeana-cased & 2 & 60\\
    SenWave-Spanish & dccuchile/bert-base-spanish-wwm-uncased & 2 & 60\\
    SST-2 & bert-base-uncased & 2 & 20\\
    \bottomrule
    \end{tabular}%
  
  }
    
      \caption{Settings of hyperparameters across datasets, all PLMs are directly loaded from the Huggingface platform.}
  \label{tab:hpm_ds}%
\end{table*}%

\begin{table*}[h!]
  \centering
    \begin{tabular}{c|cccc}
    \toprule
    Challenge  & SMP2020-EWECT & India-COVID-X & SenWave & SST-2 \\
    \midrule
    Insufficient Samples & $\checkmark$ & $\checkmark$ & $\checkmark$ & $\times$ \\
    Domain-Specific      & $\checkmark$ & $\checkmark$ & $\checkmark$ & $\times$ \\
    Imbalanced Distribution & $\checkmark$ & $\times$ & $\times$ & $\times$ \\
    Multilingual         & $\times$ & $\times$ & $\checkmark$ & $\times$ \\
    \bottomrule
    \end{tabular}
    \caption{Low-resource challenges of datasets.}
    \label{low-resource-challenge}
\end{table*}

\begin{table*}[h!]
  \centering
    \begin{tabular}{c|cccccc}
    \toprule
    Dataset & Language & \#Train & \#Test & \#Label & Avg. Length & S/D \\
    \midrule
    India-COVID-X & English & 2164  & 926   & 4     & 25.23  & 0.0127  \\
    \midrule
    SMP2020-EWECT  & Chinese & 8606  & 3000  & 6     & 54.44  & 0.1634  \\
    \midrule
    \multirow{3}[2]{*}{SenWave} & Arabic & 2210  & 553   & 6     & 26.07  & 0.1069  \\
          & Spanish & 4116  & 1029  & 6     & 19.25  & 0.1284  \\
          & French & 4116  & 1029  & 6     & 18.90  & 0.1284  \\
  \midrule
    SST-2  & English & 67000  & 18000  & 2  & 10.41  &  0.0578 \\
    \bottomrule
    \end{tabular}%
    \caption{Data statistics. S/D represents the standard deviation of label distributions.}
        \label{datastat}%
\end{table*}%

\begin{table*}[h!]
  \centering
    \begin{tabular}{c|c|cc|cc|c|c}
    \toprule
    \multirow{2}[4]{*}{Dataset} & \multirow{2}[4]{*}{Percentage} & \multicolumn{2}{c|}{DiffusionCLS} & \multicolumn{2}{c|}{PLM} & \multirow{2}[4]{*}{$\Delta F$} & \multirow{2}[4]{*}{$\Delta Acc$} \\
\cmidrule{3-6}          &       & Macro-F & Acc   & Macro-F & Acc   &       &  \\
    \midrule
    \multirow{5}[2]{*}{SMP2020-EWECT} & 0.05  & 54.93\% & 74.20\% & 54.88\% & 73.87\% & \textcolor[rgb]{ 0,  .69,  .314}{0.05\%} & \textcolor[rgb]{ 0,  .69,  .314}{0.33\%} \\
          & 0.20  & 64.35\% & 78.23\% & 63.60\% & 77.70\% & \textcolor[rgb]{ 0,  .69,  .314}{0.75\%} & \textcolor[rgb]{ 0,  .69,  .314}{0.53\%} \\
          & 0.35  & 64.49\% & 78.23\% & 63.65\% & 78.00\% & \textcolor[rgb]{ 0,  .69,  .314}{0.84\%} & \textcolor[rgb]{ 0,  .69,  .314}{0.23\%} \\
          & 0.50  & 65.09\% & 78.90\% & 65.09\% & 78.03\% & \textcolor[rgb]{ 0,  .69,  .314}{0.01\%} & \textcolor[rgb]{ 0,  .69,  .314}{0.87\%} \\
          & 1.00  & 67.98\% & 80.23\% & 66.14\% & 78.77\% & \textcolor[rgb]{ 0,  .69,  .314}{1.85\%} & \textcolor[rgb]{ 0,  .69,  .314}{1.47\%} \\
    \midrule
    \multirow{5}[2]{*}{India-COVID-X} & 0.05  & 46.33\% & 47.73\% & 44.89\% & 48.16\% & \textcolor[rgb]{ 0,  .69,  .314}{1.45\%} & \textcolor[rgb]{ .753,  0,  0}{-0.43\%} \\
          & 0.20  & 63.44\% & 63.17\% & 61.60\% & 61.34\% & \textcolor[rgb]{ 0,  .69,  .314}{1.84\%} & \textcolor[rgb]{ 0,  .69,  .314}{1.84\%} \\
          & 0.35  & 66.04\% & 65.98\% & 65.16\% & 65.12\% & \textcolor[rgb]{ 0,  .69,  .314}{0.87\%} & \textcolor[rgb]{ 0,  .69,  .314}{0.86\%} \\
          & 0.50  & 70.17\% & 69.98\% & 69.32\% & 69.01\% & \textcolor[rgb]{ 0,  .69,  .314}{0.84\%} & \textcolor[rgb]{ 0,  .69,  .314}{0.97\%} \\
          & 1.00  & 74.65\% & 74.41\% & 70.99\% & 70.63\% & \textcolor[rgb]{ 0,  .69,  .314}{3.66\%} & \textcolor[rgb]{ 0,  .69,  .314}{3.78\%} \\
    \bottomrule
    \end{tabular}%
      \caption{Experiment results on dataset SMP2020-EWECT and India-COVID-X with partial data mode, with the percentage column indicating how much data is used in the training process. $\Delta$Acc and $\Delta$F represent the performance variance between training with a data augmentation method and their corresponding baselines, i.e., without data augmentation methods.}
  \label{low-resource-cn-eng-appendix}%
\end{table*}%

\end{document}